\pdfoutput=1
\documentclass[10pt,twocolumn,letterpaper]{article}
\pdftrailerid{}

\usepackage{times,graphicx,amsmath,amssymb,booktabs}
\usepackage[dvipsnames]{xcolor}
\usepackage[numbers,sort&compress]{natbib}
\usepackage[format=plain,labelformat=simple,labelsep=period,font=small,compatibility=false]{caption}
\usepackage{multirow,colortbl,tabularx,placeins}
\usepackage[shortlabels,inline]{enumitem}
\definecolor{linkblue}{rgb}{0.21,0.49,0.74}
\usepackage[pagebackref,breaklinks,colorlinks,allcolors=linkblue]{hyperref}

\setlist[itemize]{noitemsep,leftmargin=*,topsep=0em}
\setlist[enumerate]{noitemsep,leftmargin=*,topsep=0em}

\font\subheadingfont=ptmb scaled 1100
\font\thirdheadingfont=ptmb scaled 1000
\makeatletter
\renewcommand{\@seccntformat}[1]{\csname the#1\endcsname.\nobreakspace}
\renewcommand{\section}{\@startsection{section}{1}{\z@}{-10pt plus -2pt minus -2pt}{7pt}{\large\bfseries}}
\renewcommand{\subsection}{\@startsection{subsection}{2}{\z@}{-8pt plus -2pt minus -2pt}{5pt}{\subheadingfont}}
\renewcommand{\subsubsection}{\@startsection{subsubsection}{3}{\z@}{-6pt plus -2pt minus -2pt}{3pt}{\thirdheadingfont}}
\renewcommand{\@maketitle}{
  \newpage\null\vskip 4pt
  \begin{center}
    \vspace*{-16pt}
    {\Large\bfseries\@title\par}
    \vspace{7pt}
    {\large\begin{tabular}[t]{c}\@author\end{tabular}\par}
    \vspace{6.09pt}
  \end{center}
}
\makeatother
\renewenvironment{abstract}{\centerline{\large\bfseries Abstract}\vspace*{12pt}\noindent\itshape\ignorespaces}{\vspace*{12pt}}

\title{ReMiC: Representation-Driven Token Pruning for 3D VLMs via Minimax Hierarchical Clustering}
\newcommand{\inst}[1]{\textsuperscript{#1}}
\author{
  Bangfeng Yan\inst{1}\quad Chaoning Zhang\inst{1}\thanks{\begin{tabular}[t]{@{}l@{\enspace}l@{}}
    Corresponding author & \texttt{chaoningzhang1990@gmail.com}\\
    First author & \texttt{bfyan@std.uestc.edu.cn}
  \end{tabular}}\quad Yuanhao Liu\inst{1}\quad Pengcheng Zheng\inst{1}\\
  Mengyuan Li\inst{1}\quad Wei Dong\inst{2}\quad Sheng Zheng\inst{1}\quad Yang Yang\inst{1}\\[3pt]
  {\normalsize\inst{1}\,University of Electronic Science and Technology of China}\\
  {\normalsize\inst{2}\,Xi'an University of Architecture and Technology}\\[2pt]
  {\small\url{https://github.com/yrbkf/ReMiC}}
}
\date{}
\hypersetup{pdftitle={ReMiC: Representation-Driven Token Pruning for 3D VLMs via Minimax Hierarchical Clustering}}

\begin{document}
\maketitle
\AddToHookNext{cmd/footnoterule/after}{\kern.76pt}
\enlargethispage{26pt}

\begin{abstract}
3D vision-language models (3D VLMs) have demonstrated strong capabilities in 3D scene understanding and spatial reasoning.
However, their reliance on multi-view inputs produces long visual token sequences, imposing substantial computational overhead during inference.
Existing pruning methods often rely on token salience or feature diversity as proxies for representativeness, without explicitly modeling the representation gap between the original token set and the retained subset, thereby leading to suboptimal representation of the original scene.
In this paper, we propose ReMiC, a training-free, representation-driven token pruning framework based on minimax hierarchical clustering.
ReMiC treats retained tokens as prototypes for the original observations, formulating pruning as minimizing the maximum representation error under a fixed token budget.
Local prototype initialization allocates tokens according to regional representation needs, while global prototype refinement updates the selected prototypes using cross-region information to further reduce the maximum representation error.
Extensive experiments conducted on three representative 3D VLMs across five 3D scene understanding benchmarks demonstrate the effectiveness and generalization of our method.
Notably, by retaining only 10\% of visual tokens, ReMiC preserves 93.2\% of the performance of Video-3D LLM and achieves a $2.04\times$ faster inference speed.
\end{abstract}
\section{Introduction}
\label{sec:intro}

Recent 3D vision-language models (3D VLMs) have demonstrated strong performance in 3D scene understanding, showing promise for applications in autonomous driving and embodied intelligence~\cite{zheng2025video3dllm,li2026spacedrive,wang2025omnidrive,huang2024leo,hong2023_3dllm}.
Given the scarcity of annotated 3D data, 3D VLMs~\cite{fan2026vlm3r,lin2026qwen3d,linghu2026threedrft,yu2026stream3dvlm,zheng2025_vgllm} usually build upon pretrained video-language models to leverage visual priors learned from large-scale 2D data, using multi-view images as visual inputs.
However, processing these inputs introduces a major computational bottleneck: the number of visual tokens grows linearly with the number of input views.
For example, Video-3D LLM~\cite{zheng2025video3dllm} encodes 32 input views into 6,272 visual tokens, substantially increasing the cost of LLM inference.
Visual token pruning is therefore essential for reducing this computational burden and enabling efficient inference in 3D VLMs.

\begin{figure}[t]
    \centering
    \includegraphics[width=\linewidth]{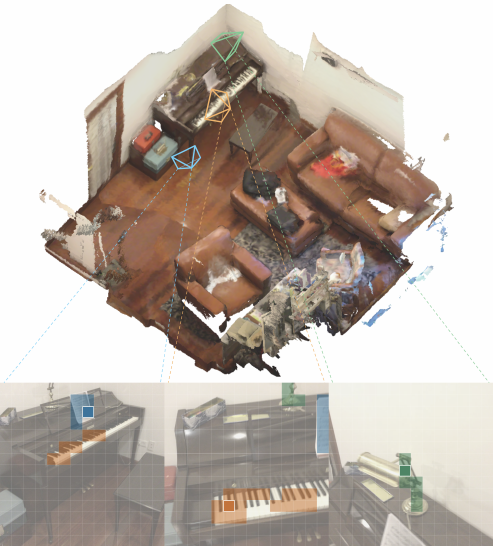}
    \caption{\textbf{Illustration of the relationship between original tokens and retained prototypes.}
    Retained prototypes are shown as solid squares with white borders. Translucent patches of the same color indicate the original tokens they represent across views.}
    \label{fig:intro_overview}
\end{figure}

Visual token pruning aims to retain a compact subset that faithfully represents the original visual input.
Attention-based methods~\cite{chen2024_fastv,yang2025visionzip,pan2026_merge3d,zhang2025vispruner,zhang2025sparsevlm} estimate token salience from attention scores and prioritize highly attended tokens, while diversity-based methods~\cite{alvar2025divprune,zhang2025cdpruner,wen2025dart,baek2026threedzip,huang2025dtc} select tokens that are dissimilar in feature space to reduce redundancy.
Both approaches rely on proxies for representativeness, without directly evaluating how well the retained subset represents the original observations.
In multi-view 3D scenes, attention ranking can overlook cross-view redundancy and concentrate the token budget on repeated observations of a few salient regions, leaving other regions underrepresented~\cite{li2026_geo3dpruner}.
Maximizing feature separation can favor outliers in feature space while discarding representative tokens that capture dominant visual patterns~\cite{ling2026coverprune}.
Such selection biases leave parts of the scene poorly represented, limiting the visual evidence available for object localization and spatial reasoning.

To address this issue, we explicitly model the representation gap between the original token set and the retained subset.
We treat the retained tokens as prototypes, with every original token assigned to one of them.
As illustrated in Fig.~\ref{fig:intro_overview}, observations of the same local scene region across views are represented by a shared retained prototype.
Each token's representation error is defined as the joint feature-spatial distance to its prototype, capturing differences in both visual content and 3D position.
To keep all original tokens close to their prototypes, we aim to minimize the maximum representation error under a fixed token budget.

Building on this formulation, we propose \textbf{ReMiC}, a training-free, two-stage framework for visual token pruning via minimax hierarchical clustering~\cite{bien2011minimax}.
\textit{Local prototype initialization} builds a minimax tree independently in each spatial region to provide candidate representations at different granularities.
These trees are jointly cut under a shared token budget to minimize the maximum representation error, retaining finer representations where coarser grouping incurs larger errors.
\textit{Global prototype refinement} then updates the selected prototypes using cross-region information to further reduce the maximum representation error.

Extensive experiments have been conducted on five 3D scene understanding benchmarks~\cite{chen2020scanrefer,zhang2023multi3drefer,chen2021scan2cap,azuma2022scanqa,ma2023sqa3d} and three representative 3D VLMs~\cite{zheng2025video3dllm,wang2025ross3d,chen2026gsreasoner}.
ReMiC consistently achieves the best average performance among existing token pruning methods across all evaluated token budgets.
Notably, with only 10\% of the visual tokens retained, ReMiC preserves 93.2\% of the original performance on Video-3D LLM, while achieving a $6.72\times$ prefill speedup and a $2.04\times$ end-to-end inference speedup.

Our contributions are summarized as threefold:

\begin{itemize}
  \item We explicitly model the representation gap between the original and retained token sets, and formulate visual token pruning as minimizing the maximum representation error under a fixed token budget.
  \item We propose ReMiC, a training-free, two-stage framework for visual token pruning via minimax hierarchical clustering.
\textit{Local prototype initialization} adaptively allocates prototypes across scene regions, while \textit{global prototype refinement} improves prototype selection by incorporating cross-region information.
  \item Extensive experiments across three 3D VLM backbones and five scene understanding benchmarks show that ReMiC maintains strong task performance under aggressive pruning while substantially accelerating inference.
\end{itemize}
\begin{figure*}[t]
    \centering
    \includegraphics[width=\textwidth]{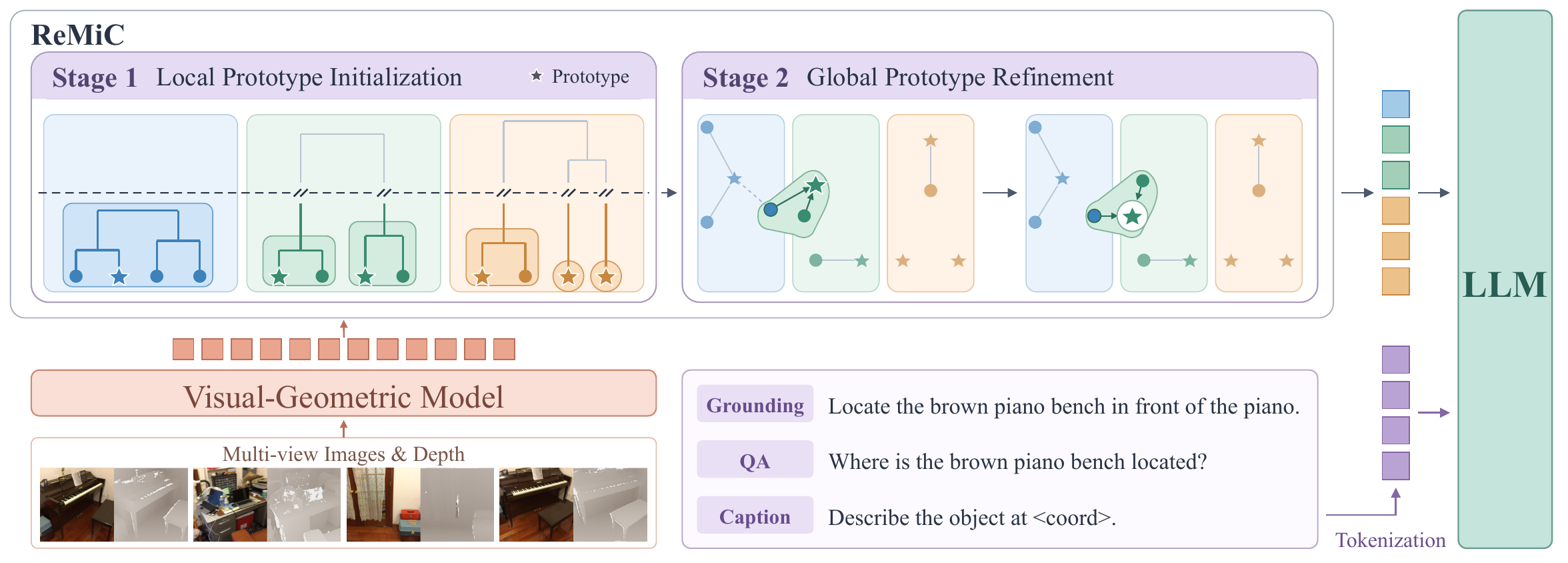}
    \caption{\textbf{Overview of the ReMiC framework.}
    ReMiC serves as a training-free, plug-and-play module inserted between the visual-geometric encoder and the LLM.
    In Stage 1, \textit{local prototype initialization} builds a minimax tree within each spatial region and jointly cuts the trees under a shared token budget to select initial prototypes.
    In Stage 2, \textit{global prototype refinement} reassigns all original tokens to their nearest prototypes across regions and reselects a prototype for each cluster.}
    \label{fig:method_overview}
\end{figure*}
\section{Related Work}
\label{sec:related_work}

\subsection{3D Vision-Language Models}
Recent advances in vision-language models have spurred growing interest in extending their capabilities to 3D scene understanding~\cite{hong2023_3dllm,huang2024leo,zhi2025lscenellm,deng2025threedllava}.
Early approaches equip language models with 3D understanding through point-cloud encoding or object-centric scene representations~\cite{xu2024pointllm,qi2024gpt4point,chen2023_ll3da,huang2023_chatscene}.
Given the scarcity of 3D--language paired data, a growing line of work~\cite{fan2026vlm3r,linghu2026threedrft,lin2026qwen3d,yu2026stream3dvlm} has explored adapting pretrained video-language models for 3D scene understanding, thereby leveraging rich visual priors learned from large-scale 2D data.
For example, Video-3D LLM~\cite{zheng2025video3dllm} and LLaVA-3D~\cite{zhu2024_llava3d} incorporate 3D positional embeddings into multi-view visual features, grounding observations across views in a shared 3D coordinate system.
To further enhance spatial understanding, VG-LLM~\cite{zheng2025_vgllm} and GS-Reasoner~\cite{chen2026gsreasoner} incorporate geometric features from pretrained 3D encoders into visual representations.
Meanwhile, Ross3D~\cite{wang2025ross3d} introduces cross-view and global-view reconstruction objectives to provide 3D-aware visual supervision during training.
Despite the rapid progress of 3D VLMs, they still face substantial inference overhead due to the long visual token sequences generated by multi-view inputs.

\subsection{Visual Token Pruning}
Visual token pruning offers an effective way to reduce VLM inference costs by compressing visual sequences while preserving essential information~\cite{zhang2026rcp,zheng2026causalprune,cao2026tprl}.
Existing methods can be broadly categorized into two groups.
The first is attention-based pruning, which prioritizes visual tokens with high attention scores.
For example, FastV~\cite{chen2024_fastv} uses attention scores from an early LLM decoder layer to remove low-scoring visual tokens from subsequent layers, while VisionZip~\cite{yang2025visionzip} uses vision encoder attention to select dominant tokens and merges the remaining tokens into contextual tokens based on feature similarity.
The second is diversity-based pruning, which seeks to reduce redundancy in the retained set by selecting visual tokens that are dissimilar in feature space.
In this category, DivPrune~\cite{alvar2025divprune} aims to maximize the minimum pairwise feature distance among retained tokens, while DART~\cite{wen2025dart} selects a small set of visual and text tokens as pivots and retains visual tokens with low cosine similarity to these pivots.

Recent work extends visual token pruning to multi-view 3D scenes by incorporating geometric information to guide token selection across views.
DTC~\cite{huang2025dtc} removes redundant tokens through feature matching within progressively enlarged voxels.
Geo3DPruner~\cite{li2026_geo3dpruner} uses geometry-aware global attention to guide intra- and inter-voxel selection, while Merge3D~\cite{pan2026_merge3d} selects dominant tokens using 2D attention and merges the remaining tokens based on semantic and geometric similarity.
\section{Method}
\label{sec:method}

We first describe visual token construction in 3D VLMs (Sec.~\ref{sec:preliminaries}) and formulate visual token pruning via minimax hierarchical clustering (Sec.~\ref{sec:minimax_formulation}).
Building on this formulation, we present ReMiC, a training-free, two-stage framework for visual token pruning that aims to minimize the maximum representation error across the scene (Fig.~\ref{fig:method_overview}).
\textit{Local prototype initialization} (Sec.~\ref{sec:local_initialization}) adaptively allocates tokens across scene regions according to their representation needs.
\textit{Global prototype refinement} (Sec.~\ref{sec:global_refinement}) then updates the selected prototypes using cross-region information to further reduce the maximum representation error.

\subsection{Preliminaries}
\label{sec:preliminaries}

We consider multi-view 3D VLMs that take RGB images, depth maps, and camera parameters as inputs to construct spatially grounded visual tokens for LLM inference.
The 3D VLM extracts patch-level features from RGB images using a vision encoder, projects them into the LLM embedding space, and applies spatial pooling to obtain $N$ visual features $f_i\in\mathbb R^d$.
Using depth maps and camera parameters, the model maps visual observations into a shared world coordinate system, assigning each token a 3D coordinate $x_i\in\mathbb R^3$.
The backbone then applies a spatial encoding function $\phi(\cdot)$ to obtain the spatial feature $s_i=\phi(x_i)\in\mathbb R^d$, which is combined with the visual feature to yield the fused token representation $z_i=f_i+s_i\in\mathbb R^d$.

We represent each 3D visual token as $v_i=(z_i,x_i)$, forming the scene token set $\mathcal V=\{v_i\}_{i=1}^{N}$.
Given a token budget $B\ll N$, ReMiC selects a subset $\mathcal V'\subseteq\mathcal V$ with $|\mathcal V'|=B$.
The retained visual token embeddings are stacked into $\mathbf Z\in\mathbb R^{B\times d}$ and concatenated with textual tokens, including system tokens $T_{\mathrm{system}}$ and query tokens $T_{\mathrm{query}}$, to form the LLM input:
\begin{equation}
    X=\bigl[T_{\mathrm{system}};\mathbf Z;T_{\mathrm{query}}\bigr]
    \in\mathbb R^{(B+L_{\mathrm{text}})\times d}.
    \label{eq:llm_input}
\end{equation}

\subsection{Minimax Formulation for Token Pruning}
\label{sec:minimax_formulation}

We formulate visual token pruning as minimax hierarchical clustering~\cite{bien2011minimax}, with each retained token serving as a prototype for a cluster of original tokens.
We aim to minimize the maximum representation error across these clusters.

\paragraph{Joint Feature--Spatial Distance.}
To preserve both visual content and spatial structure during pruning, we define a joint distance for minimax hierarchical clustering.
Specifically, we measure feature dissimilarity with cosine distance and spatial separation with Euclidean distance:
\begin{equation}
    d_f(v_i,v_j)\!=\!1-\cos(z_i,z_j),\;d_x(v_i,v_j)\!=\!\|x_i-x_j\|_2.
    \label{eq:token_distances}
\end{equation}
We normalize each distance component by its maximum over all input token pairs, yielding $\hat{d}_f$ and $\hat{d}_x$.
The normalized spatial distance $\hat{d}_x$ is then logarithmically transformed to emphasize short-range spatial differences~\cite{ling2026coverprune}.
We combine the resulting components as
\begin{equation}
    d(v_i,v_j)=\lambda\hat d_f(v_i,v_j)+(1-\lambda)\log_2\!\bigl(1+\hat d_x(v_i,v_j)\bigr),
    \label{eq:joint_distance}
\end{equation}
where $\lambda\in[0,1]$ balances feature and spatial distances.

\paragraph{Minimax Prototype Selection.}
To control the maximum representation error across all original tokens in a cluster $C\subseteq\mathcal V$, a candidate token $v\in C$ is evaluated by its distance to the farthest token in $C$:
\begin{equation}
    d_{\max}(v,C)=\max_{v'\in C}d(v,v').
    \label{eq:minimax_max_distance}
\end{equation}
The minimax prototype $p_C$ minimizes $d_{\max}(v,C)$ over $v\in C$, and the minimum value of $d_{\max}(v,C)$ is the minimax radius $r(C)$~\cite{bien2011minimax}:
\begin{equation}
    r(C)=\min_{v\in C}d_{\max}(v,C),\;p_C\in\operatorname*{arg\,min}_{v\in C}d_{\max}(v,C).
    \label{eq:minimax_prototype}
\end{equation}
The radius $r(C)$ measures the maximum representation error when $p_C$ represents all original tokens in $C$.
The minimax linkage between two clusters $G$ and $H$ is defined as the radius of their union:
\begin{equation}
    d(G,H)=r(G\cup H).
    \label{eq:minimax_linkage}
\end{equation}
Starting from $N$ singleton clusters, minimax hierarchical clustering repeatedly merges the pair with the smallest linkage, yielding a prototype and an associated radius for each newly formed cluster.

\paragraph{Token Pruning Objective.}
Given a token budget $B$, we cut the minimax tree to obtain $B$ disjoint clusters that collectively cover all original tokens.
Let $\mathcal C$ denote these clusters, each represented by its prototype $p_C$.
The largest cluster radius measures the maximum representation error across all original tokens relative to their respective cluster prototypes:
\begin{equation}
    R(\mathcal C)=\max_{C\in\mathcal C}r(C).
    \label{eq:pruning_objective}
\end{equation}
Intuitively, $R(\mathcal C)$ asks whether the prototypes adequately represent even the worst-represented parts of the scene, providing a uniform upper bound on the representation error of every original token.
Keeping this bound small therefore guarantees that all scene regions are well represented.
To achieve low error, a few prototypes may suffice for regions with limited variation, but regions with greater spatial or feature variation may require more.
Reducing $R(\mathcal C)$ thus favors allocating more prototypes to regions that require finer representation, while allowing simpler regions to be represented with fewer.
This promotes spatial coverage while adapting representation granularity to local scene complexity.
Given these properties, we aim to minimize $R(\mathcal C)$ under a fixed token budget.

\subsection{Local Prototype Initialization}
\label{sec:local_initialization}

Stage 1 constructs local minimax trees to organize candidate clusters at different granularities, then jointly cuts these trees to obtain $B$ initial prototypes.

\paragraph{Local Tree Construction.}
We sort the tokens by their Morton codes~\cite{morton1966geodetic} and evenly partition the resulting sequence into $L$ groups, where $L\le B$.
Tokens adjacent in this order tend to be proximal in 3D space, allowing contiguous segments of the sequence to serve as local scene regions.
A minimax tree is then built independently within each region as described in Sec.~\ref{sec:minimax_formulation}.
This design improves computational efficiency through parallel tree construction and balanced workloads, while preserving downstream task performance by maintaining spatial locality.

\paragraph{Adaptive Tree Cutting.}
A natural approach to initializing $B$ prototypes is to allocate equal budgets to regions and cut their trees independently.
However, regions with different levels of complexity may require different numbers of prototypes, so we jointly cut these trees to allocate the budget adaptively.
Let $\mathcal F$ denote the valid cuts of the local trees into $B$ disjoint clusters covering all original tokens.
The optimal cut satisfies
\begin{equation}
    \mathcal C^\star \in
    \operatorname*{arg\,min}_{\mathcal C\in\mathcal F}
    R(\mathcal C).
    \label{eq:adaptive_cut}
\end{equation}

By the no-inversion property of minimax trees~\cite{bien2011minimax}, we can obtain the optimal cut by sorting all local merges in ascending order of minimax linkage $d(G,H)$ and applying the first $N-B$ merges.
Consequently, regions that allow more merges at small radii retain fewer prototypes, leaving more of the shared budget for regions where further merging would produce larger representation errors.

\begin{table*}[t]
    \centering
\caption{\textbf{Performance comparison with other baselines on Video-3D LLM.}
    Avg. reports the average percentage of original performance retained across the five benchmarks and nine metrics.}
\label{tab:comparison_10_25}
\begingroup
\setlength{\tabcolsep}{4.5pt}
\renewcommand{\arraystretch}{1.18}
\resizebox{\linewidth}{!}{\begin{tabular}{@{}lc*{10}{c}@{}}
\toprule
\multirow{2}{*}{\textbf{Method}} & \multirow{2}{*}{\textbf{\shortstack{Retention\\Ratio {\boldmath$R$}}}} & \multicolumn{2}{c}{\textbf{ScanRefer}} & \multicolumn{2}{c}{\textbf{Multi3DRefer}} & \multicolumn{2}{c}{\textbf{Scan2Cap}} & \multicolumn{2}{c}{\textbf{ScanQA}} & \textbf{SQA3D} & \textbf{Avg.} \\
\cmidrule(lr){3-4}\cmidrule(lr){5-6}\cmidrule(lr){7-8}\cmidrule(lr){9-10}\cmidrule(lr){11-11}
& & Acc@0.25 & Acc@0.5 & F1@0.25 & F1@0.5 & B-4@0.5 & C@0.5 & CIDEr & EM & EM & (\%) \\
\midrule
Video-3D LLM~\cite{zheng2025video3dllm} {\scriptsize(CVPR25)} & 100\% & 58.18 & 51.77 & 57.42 & 52.15 & 41.36 & 84.11 & 102.04 & 30.10 & 58.45 & 100.00 \\
\midrule
VisionZip~\cite{yang2025visionzip} {\scriptsize(CVPR25)} & \multirow{6}{*}{25\%} & 57.36 & 50.98 & 56.27 & 51.13 & 38.98 & 73.38 & \underline{98.41} & 29.09 & \underline{58.31} & \underline{96.38} \\
DivPrune~\cite{alvar2025divprune} {\scriptsize(CVPR25)} &  & 57.25 & 50.80 & 56.17 & 51.04 & \underline{39.22} & \underline{73.64} & 97.17 & 28.86 & 58.14 & 96.13 \\
V-CAST~\cite{lin2026vcast} {\scriptsize(NeurIPS26)} &  & 56.99 & 50.55 & 56.26 & 51.02 & 38.15 & 69.72 & 97.81 & 28.77 & 57.43 & 95.14 \\
3DZip~\cite{baek2026threedzip} {\scriptsize(ECCV26)} &  & 57.17 & 50.89 & 56.22 & 51.19 & 38.87 & 73.42 & 98.08 & 28.81 & 57.97 & 96.11 \\
CoVeR~\cite{bui2026cover} {\scriptsize(arxiv26)} &  & \underline{57.41} & \textbf{51.14} & \underline{56.37} & \underline{51.26} & 39.07 & 72.96 & 97.18 & \underline{29.13} & 58.20 & 96.31 \\
\textbf{ReMiC (Ours)} &  & \textbf{57.52} & \underline{51.13} & \textbf{56.42} & \textbf{51.29} & \textbf{39.61} & \textbf{76.61} & \textbf{98.89} & \textbf{29.41} & \textbf{58.48} & \textbf{97.31} \\
\midrule
VisionZip~\cite{yang2025visionzip} {\scriptsize(CVPR25)} & \multirow{6}{*}{20\%} & 56.68 & 50.43 & 55.88 & 50.84 & 38.38 & 69.68 & 97.25 & 28.36 & \textbf{58.03} & 94.90 \\
DivPrune~\cite{alvar2025divprune} {\scriptsize(CVPR25)} &  & \underline{57.41} & \underline{50.96} & 55.96 & \underline{50.87} & \textbf{38.95} & \underline{73.26} & 96.91 & 28.66 & \underline{57.74} & \underline{95.82} \\
V-CAST~\cite{lin2026vcast} {\scriptsize(NeurIPS26)} &  & 56.93 & 50.53 & \underline{55.98} & 50.84 & 37.36 & 64.60 & \underline{97.37} & 28.75 & 57.12 & 94.03 \\
3DZip~\cite{baek2026threedzip} {\scriptsize(ECCV26)} &  & 56.94 & 50.66 & 55.93 & \textbf{50.90} & 38.59 & 72.76 & 97.36 & 28.56 & 57.72 & 95.51 \\
CoVeR~\cite{bui2026cover} {\scriptsize(arxiv26)} &  & 57.12 & 50.78 & 55.88 & 50.80 & 38.92 & 72.88 & 96.16 & \underline{28.77} & 57.37 & 95.53 \\
\textbf{ReMiC (Ours)} &  & \textbf{57.66} & \textbf{51.28} & \textbf{55.99} & \underline{50.87} & \underline{38.94} & \textbf{74.06} & \textbf{98.12} & \textbf{29.30} & 57.66 & \textbf{96.40} \\
\midrule
VisionZip~\cite{yang2025visionzip} {\scriptsize(CVPR25)} & \multirow{6}{*}{15\%} & 56.41 & 50.04 & 54.76 & 49.82 & 37.63 & 66.94 & 93.95 & 27.59 & 57.15 & 92.96 \\
DivPrune~\cite{alvar2025divprune} {\scriptsize(CVPR25)} &  & \underline{56.71} & 50.31 & 55.32 & 50.33 & \underline{38.00} & 68.86 & 94.80 & 28.26 & 56.81 & 93.91 \\
V-CAST~\cite{lin2026vcast} {\scriptsize(NeurIPS26)} &  & 56.53 & 50.16 & 55.24 & 50.19 & 35.76 & 55.99 & \underline{96.43} & \underline{28.47} & 57.23 & 91.84 \\
3DZip~\cite{baek2026threedzip} {\scriptsize(ECCV26)} &  & 56.64 & \underline{50.32} & \underline{55.39} & \underline{50.42} & 37.87 & \underline{69.33} & 96.19 & 28.34 & 57.32 & \underline{94.24} \\
CoVeR~\cite{bui2026cover} {\scriptsize(arxiv26)} &  & 56.48 & 50.05 & 55.31 & 50.28 & 37.79 & 68.75 & 94.25 & 28.09 & \underline{57.66} & 93.77 \\
\textbf{ReMiC (Ours)} &  & \textbf{56.84} & \textbf{50.58} & \textbf{55.45} & \textbf{50.45} & \textbf{38.32} & \textbf{69.82} & \textbf{98.02} & \textbf{29.09} & \textbf{57.74} & \textbf{95.10} \\
\midrule
VisionZip~\cite{yang2025visionzip} {\scriptsize(CVPR25)} & \multirow{6}{*}{10\%} & 55.96 & 49.73 & 52.75 & 48.14 & 36.57 & 61.59 & 90.96 & 27.06 & 55.30 & 90.19 \\
DivPrune~\cite{alvar2025divprune} {\scriptsize(CVPR25)} &  & 55.97 & 49.63 & 54.12 & 49.32 & 36.91 & 62.70 & 92.18 & 27.14 & \underline{56.83} & 91.38 \\
V-CAST~\cite{lin2026vcast} {\scriptsize(NeurIPS26)} &  & 55.99 & 49.73 & 54.19 & 49.27 & 34.39 & 48.46 & 94.10 & \underline{27.91} & 56.27 & 89.24 \\
3DZip~\cite{baek2026threedzip} {\scriptsize(ECCV26)} &  & \underline{56.30} & \underline{49.92} & \textbf{54.75} & \textbf{49.82} & \underline{37.45} & \underline{66.80} & \underline{94.78} & 27.83 & 56.69 & \underline{92.93} \\
CoVeR~\cite{bui2026cover} {\scriptsize(arxiv26)} &  & 56.11 & 49.71 & 53.33 & 48.43 & 36.52 & 61.70 & 91.45 & 27.66 & 56.64 & 90.92 \\
\textbf{ReMiC (Ours)} &  & \textbf{56.36} & \textbf{49.95} & \underline{54.59} & \underline{49.61} & \textbf{37.71} & \textbf{66.82} & \textbf{95.04} & \textbf{28.32} & \textbf{56.86} & \textbf{93.19} \\
\bottomrule
\end{tabular}}
\endgroup
\end{table*}

\subsection{Global Prototype Refinement}
\label{sec:global_refinement}

The regional partitioning in Stage 1 confines each prototype to representing tokens within its own region.
However, tokens in one region may be better represented by a prototype in another.
Stage 2 therefore reorganizes clusters globally and reselects their prototypes to further reduce the maximum representation error across the scene.

\paragraph{Nearest-Prototype Reassignment.}
Each original token $v$ is reassigned to the nearest of the $B$ initial prototypes under the joint distance $d$:
\begin{equation}
    a(v)\in\operatorname*{arg\,min}_{C\in\mathcal C}d(v,p_C).
    \label{eq:global_reassignment}
\end{equation}
The tokens assigned to $p_C$ form the cluster $\widetilde C=\{v\in\mathcal V\mid a(v)=C\}$, with $a(p_C)=C$.
This reassignment updates the set of tokens each prototype needs to represent, allowing its subsequent reselection to account for members from across the scene rather than only those in its initial cluster.

\paragraph{Constrained Prototype Reselection.}
For each reassigned cluster $\widetilde C$, we reselect its prototype from $C\cap\widetilde C$ to minimize the maximum distance to all tokens in $\widetilde C$:
\begin{equation}
    p'_C\in\operatorname*{arg\,min}_{v\in C\cap\widetilde C}
    d_{\max}(v,\widetilde C).
    \label{eq:prototype_reselection}
\end{equation}
The intersection constraint retains one prototype per initial cluster, preserving the token budget allocated to each region in Stage 1.
We perform a single round of reassignment and prototype reselection.
The reselected prototypes form the final visual token set $\mathcal V'=\{p'_C\mid C\in\mathcal C\}$, better matching representation needs across the scene.
\section{Experiments}
\label{sec:experiments}

\begin{table*}[t]
    \centering
    \caption{\textbf{Performance comparison with other baselines on Ross3D and GS-Reasoner.}
    Avg. reports the average percentage of original performance retained across the three benchmarks and ten metrics.}
    \label{tab:backbone_comparison_preview}
    \small
    \setlength{\tabcolsep}{2pt}
    \renewcommand{\arraystretch}{1.08}
    \begin{tabular*}{\linewidth}{@{\extracolsep{\fill}}lc*{11}{c}@{}}
        \toprule
        \multirow{2}{*}{\textbf{Method}} &
        \multirow{2}{*}{\textbf{\shortstack{Retention\\ratio {\boldmath$R$}}}} &
        \multicolumn{4}{c}{\textbf{Scan2Cap}} &
        \multicolumn{5}{c}{\textbf{ScanQA}} &
        \textbf{SQA3D} & \textbf{Avg.} \\
        \cmidrule(lr){3-6}\cmidrule(lr){7-11}\cmidrule(lr){12-12}
        & & B-4 & ROUGE-L & CIDEr & METEOR
        & B-4 & ROUGE-L & CIDEr & METEOR & EM
        & EM & (\%) \\
        \midrule
        \rowcolor{black!8}[0pt][0pt]
        \multicolumn{13}{@{}c@{}}{\textit{Ross3D}} \\
        Vanilla & 100\% & 40.62 & 61.87 & 77.83 & 28.41 & 18.06 & 50.74 & 107.17 & 20.83 & 30.80 & 62.97 & 100.00 \\
        \midrule
        CoVeR & \multirow{4}{*}{20\%}
            & \underline{38.17} & 60.84 & 68.44 & 27.32
            & \underline{16.75} & \underline{48.61} & \underline{101.06} & \underline{19.88} & \underline{30.01}
            & \textbf{61.35} & \underline{94.95} \\
        3DZip &
            & \textbf{38.77} & \textbf{61.07} & \textbf{69.94} & \textbf{27.55}
            & 15.94 & 47.72 & 98.81 & 19.44 & 29.33
            & 60.73 & 94.04 \\
        V-CAST &
            & 35.54 & 59.93 & 57.77 & 26.53
            & 16.51 & 48.35 & 100.28 & 19.54 & 29.58
            & 60.56 & 91.83 \\
        \textbf{ReMiC (Ours)} &
            & 37.84 & \underline{60.88} & \underline{68.49} & \underline{27.40}
            & \textbf{17.13} & \textbf{48.91} & \textbf{102.06} & \textbf{20.03} & \textbf{30.40}
            & \underline{61.07} & \textbf{95.42} \\
        \midrule
        CoVeR & \multirow{4}{*}{10\%}
            & 36.66 & 60.45 & 62.02 & 26.83
            & 15.61 & 46.74 & 96.04 & 19.01 & \underline{28.62}
            & \underline{59.70} & 90.92 \\
        3DZip &
            & \underline{37.15} & \textbf{60.56} & \underline{63.87} & \textbf{26.97}
            & 15.72 & \underline{47.05} & \underline{96.63} & \underline{19.10} & 28.49
            & \textbf{59.87} & \underline{91.55} \\
        V-CAST &
            & 31.98 & 58.43 & 43.84 & 25.10
            & \textbf{16.25} & 47.03 & 96.41 & 18.95 & 28.45
            & 59.45 & 86.82 \\
        \textbf{ReMiC (Ours)} &
            & \textbf{37.27} & \underline{60.48} & \textbf{63.93} & \underline{26.95}
            & \underline{16.11} & \textbf{47.38} & \textbf{97.67} & \textbf{19.29} & \textbf{29.07}
            & 58.91 & \textbf{92.07} \\
        \midrule
        \rowcolor{black!8}[0pt][0pt]
        \multicolumn{13}{@{}c@{}}{\textit{GS-Reasoner}} \\
        Vanilla & 100\% & 51.27 & 71.61 & 105.68 & 32.71 & 17.82 & 50.50 & 107.30 & 20.77 & 30.33 & 60.87 & 100.00 \\
        \midrule
        CoVeR & \multirow{4}{*}{20\%}
            & \textbf{49.58} & \textbf{70.93} & \underline{96.41} & \textbf{31.99}
            & 16.42 & \underline{48.06} & 100.27 & \underline{19.62} & \underline{28.58}
            & \underline{58.77} & \underline{95.08} \\
        3DZip &
            & 48.22 & 70.31 & 94.01 & 31.60
            & 16.21 & 47.75 & 99.11 & 19.21 & \underline{28.58}
            & 57.09 & 93.62 \\
        V-CAST &
            & 49.00 & 70.72 & 93.11 & 31.68
            & \textbf{16.83} & \textbf{48.47} & \textbf{101.30} & \textbf{19.77} & \textbf{28.79}
            & 58.54 & 95.04 \\
        \textbf{ReMiC (Ours)} &
            & \underline{49.38} & \underline{70.92} & \textbf{97.19} & \underline{31.95}
            & \underline{16.57} & 48.02 & \underline{100.96} & 19.57 & 28.47
            & \textbf{58.91} & \textbf{95.21} \\
        \midrule
        CoVeR & \multirow{4}{*}{10\%}
            & \underline{47.89} & \underline{70.21} & \underline{90.07} & \textbf{31.30}
            & 15.50 & 46.13 & 95.57 & 18.85 & 27.57
            & \underline{57.18} & \underline{91.54} \\
        3DZip &
            & 46.91 & 69.83 & 86.60 & 30.98
            & 16.29 & 46.48 & 95.95 & 18.78 & 27.47
            & 56.01 & 91.16 \\
        V-CAST &
            & 45.44 & 69.25 & 76.28 & 30.14
            & \underline{16.68} & \underline{46.71} & \underline{97.66} & \underline{19.08} & \underline{27.83}
            & \textbf{57.46} & 90.48 \\
        \textbf{ReMiC (Ours)} &
            & \textbf{48.08} & \textbf{70.26} & \textbf{90.26} & \underline{31.29}
            & \textbf{16.69} & \textbf{46.91} & \textbf{97.78} & \textbf{19.14} & \textbf{28.02}
            & 56.27 & \textbf{92.76} \\
        \bottomrule
    \end{tabular*}
\end{table*}

\subsection{Experimental Settings}
\label{sec:experimental_settings}

\paragraph{Benchmarks.}
We evaluate ReMiC on five 3D scene understanding benchmarks: ScanRefer~\cite{chen2020scanrefer} for single-object visual grounding, Multi3DRefer~\cite{zhang2023multi3drefer} for multi-target visual grounding, Scan2Cap~\cite{chen2021scan2cap} for 3D dense captioning, ScanQA~\cite{azuma2022scanqa} for 3D spatial understanding, and SQA3D~\cite{ma2023sqa3d} for situated reasoning conditioned on a specified position and orientation.
Together, these benchmarks cover three key 3D scene understanding tasks: object localization, scene description, and spatial reasoning.
We follow the standard evaluation settings and metrics for each benchmark.

\paragraph{Baselines.}
We compare ReMiC with five state-of-the-art visual token pruning methods:
\mbox{1) VisionZip}~\cite{yang2025visionzip}, which selects dominant tokens using attention scores and merges the remaining tokens into contextual tokens;
\mbox{2) DivPrune}~\cite{alvar2025divprune}, which formulates token selection as a max-min diversity problem (MMDP)~\cite{porumbel2011maxmin};
\mbox{3) V-CAST}~\cite{lin2026vcast}, which allocates frame budgets using temporal curvature and selects tokens with dual spatial anchors;
\mbox{4) 3DZip}~\cite{baek2026threedzip}, which combines voxel aggregation, diversity-based anchor selection, and spatially constrained token merging; and
\mbox{5) CoVeR}~\cite{bui2026cover}, which selects tokens based solely on 3D coordinates to maximize spatial coverage.

\paragraph{Implementation Details.}
We evaluate ReMiC on three representative 3D VLMs: Video-3D LLM~\cite{zheng2025video3dllm}, Ross3D~\cite{wang2025ross3d}, and GS-Reasoner~\cite{chen2026gsreasoner}, which cover diverse architectures to ensure generality.
Following the official settings, all three models uniformly sample 32 frames per scene and adopt the grid-based newline token format~\cite{zhang2025llavavideo}.
ReMiC operates on visual tokens immediately before the LLM prefill stage without modifying the backbone.
Following FastVID~\cite{shen2025fastvid}, we retain newline tokens only for nonempty rows after pruning.
Unless otherwise specified, we set the number of local regions to $L=196$ and the joint distance weight to $\lambda=0.5$.
All ablation studies are conducted on Video-3D LLM at 10\% token retention.
We use NVIDIA RTX 3090 GPUs for all experiments.

\subsection{Main Comparison}
\label{sec:main_results}

\paragraph{Comparison on Video-3D LLM.}
As shown in Table~\ref{tab:comparison_10_25}, ReMiC achieves the best aggregate performance at every token budget.
This consistency aligns with our minimax tree design, which adapts representation granularity to the available token budget.
In contrast, 3DZip is less competitive at higher token budgets, suggesting that early voxel aggregation may prematurely discard fine-grained details that could otherwise be preserved.
At 10\% token retention, ReMiC retains 93.19\% of the original performance on average, compared with 90.19\% for \mbox{VisionZip} and 91.38\% for \mbox{DivPrune}.
These results suggest that selecting prototypes to represent the original token set preserves scene information more effectively under aggressive pruning than prioritizing token importance or feature diversity.
ReMiC also consistently outperforms CoVeR on average, supporting the value of minimizing representation error in both feature and spatial domains rather than optimizing spatial coverage alone.

\paragraph{Comparison on Ross3D and GS-Reasoner.}
Table~\ref{tab:backbone_comparison_preview} compares ReMiC with baseline methods on Ross3D and GS-Reasoner, showing that ReMiC consistently achieves the best average performance across both backbones and token budgets.
Notably, these two backbones employ different mechanisms to incorporate 3D spatial information, demonstrating that ReMiC can adapt to diverse 3D feature representations.
In contrast, V-CAST exhibits substantial performance variations across the two backbones, highlighting the limitations of video-oriented pruning in handling 3D spatial features and the need for methods specifically designed for 3D VLMs.
At $R = 10\%$, ReMiC retains 92.07\% and 92.76\% of the original performance on Ross3D and GS-Reasoner, outperforming the second-best methods by 0.52 and 1.22 percentage points, respectively.
These results demonstrate that minimizing the maximum representation error provides an effective pruning strategy across 3D VLM backbones with different spatial representations.

\begingroup
\setlength{\intextsep}{\medskipamount}
\subsection{Ablation Studies}
\label{sec:ablation_studies}

\paragraph{Ablation on Region Partitioning.}
As illustrated in Tab.~\ref{tab:ablation_groups}, we conduct an ablation study on the region count $L$, which controls the granularity of spatial partitioning.
In particular, $L=1$ denotes global minimax clustering without region partitioning.
The results show that region partitioning improves both pruning efficiency and task performance by enabling parallel tree construction and exploiting spatial locality.
Notably, pruning time remains nearly constant for $L$ ranging from 128 to 392, as further partitioning yields diminishing savings in local tree construction while other pruning costs remain largely unchanged.
Peak performance is achieved at $L=196$, suggesting that overly coarse partitions weaken spatial locality, whereas overly fine partitions split tokens from the same object across regions.

\begin{table}[hbp]
    \centering
    \caption{\textbf{Ablation study on region count {\boldmath$L$}.}}
    \label{tab:ablation_groups}
    \begingroup
    \footnotesize
    \setlength{\tabcolsep}{1pt}
    \renewcommand{\arraystretch}{1.18}
    \begin{tabular*}{\linewidth}{@{\extracolsep{\fill}}c*{8}{c}@{}}
        \toprule
        \multirow{2}{*}{\boldmath$L$} & \multicolumn{2}{c}{\textbf{ScanRef}} & \multicolumn{2}{c}{\textbf{ScanCap}} & \multicolumn{2}{c}{\textbf{ScanQA}} & \textbf{Avg.} & \textbf{Pruning} \\
        \cmidrule(lr){2-3}\cmidrule(lr){4-5}\cmidrule(lr){6-7}
        & Acc@0.25 & Acc@0.5 & B-4@0.5 & C@0.5 & CIDEr & EM & (\%) & (s) \\
        \midrule
        1 & 56.31 & 50.02 & 37.28 & 65.13 & 93.68 & 27.94 & 90.94 & 0.434 \\
        64 & \textbf{56.45} & \textbf{50.17} & 37.51 & 65.21 & 94.32 & 27.81 & 91.16 & 0.181 \\
        128 & \textbf{56.45} & 50.13 & 37.60 & 66.05 & 93.97 & 27.87 & 91.33 & \textbf{0.098} \\
        196 & 56.36 & 49.95 & \textbf{37.71} & \textbf{66.82} & \textbf{95.04} & \textbf{28.32} & \textbf{91.87} & 0.099 \\
        392 & 56.17 & 49.89 & 37.19 & 65.61 & 93.34 & 27.74 & 90.75 & 0.101 \\
        \bottomrule
    \end{tabular*}
    \endgroup
\end{table}

\paragraph{Ablation on Tree Cutting Strategies.}
As shown in Tab.~\ref{tab:regional_budgets}, we compare our adaptive tree cutting strategy (Ada.) with two variants: uniform allocation (Unif.), which distributes the token budget evenly across regions, and shuffled allocation (Shuf.), which randomly permutes the regional budgets produced by Ada.
Ada.\ achieves the best overall performance (91.87\%), compared with 91.54\% for Unif.\ and 91.10\% for Shuf.
Unif.\ overlooks differences in regional representation needs, allocating more tokens than necessary to simpler regions while leaving complex regions underrepresented.
Meanwhile, Shuf.\ preserves the same set of regional budgets as Ada.\ but breaks their correspondence with regional representation needs, resulting in even lower performance than Unif.
Together, these results demonstrate the importance of adapting token allocation to regional representation needs.

\begin{table}[htbp]
    \centering
    \caption{\textbf{Ablation Study on Tree Cutting Strategies.}}
    \label{tab:regional_budgets}
    \label{fig:regional_budgets}
    \begingroup
    \footnotesize
    \setlength{\tabcolsep}{1pt}
    \renewcommand{\arraystretch}{1.18}
    \begin{tabular*}{\linewidth}{@{\extracolsep{\fill}}l*{7}{c}@{}}
        \toprule
        \multirow{2}{*}{\textbf{Setting}} & \multicolumn{2}{c}{\textbf{ScanRef}} & \multicolumn{2}{c}{\textbf{ScanCap}} & \multicolumn{2}{c}{\textbf{ScanQA}} & \textbf{Avg.} \\
        \cmidrule(lr){2-3}\cmidrule(lr){4-5}\cmidrule(lr){6-7}
        & Acc@0.25 & Acc@0.5 & B-4@0.5 & C@0.5 & CIDEr & EM & (\%) \\
        \midrule
        Unif. & \textbf{56.36} & 49.86 & 37.41 & \textbf{66.82} & 93.99 & 28.30 & 91.54 \\
        Shuf. & 56.28 & 49.85 & 37.30 & 65.42 & 93.92 & 28.15 & 91.10 \\
        \midrule
        \textbf{Ada.} & \textbf{56.36} & \textbf{49.95} & \textbf{37.71} & \textbf{66.82} & \textbf{95.04} & \textbf{28.32} & \textbf{91.87} \\
        \bottomrule
    \end{tabular*}
    \endgroup
\end{table}

\paragraph{Ablation on Prototype Refinement Strategies.}
As shown in Tab.~\ref{tab:ablation_refinement}, we compare our constrained refinement strategy (w/ All) with two variants: no refinement (w/o Refine.), which retains the prototypes selected in Stage~1, and unconstrained refinement (w/o Const.), which removes the candidate constraint during prototype reselection.
Our full method achieves the best overall performance (91.87\%), compared with 91.40\% for w/o Refine.\ and 91.27\% for w/o Const.
Without refinement, the initial spatial partition restricts each token's assignment to prototypes within its own region, even when more suitable prototypes exist elsewhere in the scene.
Meanwhile, removing the candidate constraint allows prototypes of different clusters to concentrate in the same area, weakening spatial coverage and resulting in even lower performance than no refinement.
Together, these results demonstrate that constrained refinement improves representation quality while preserving scene-level coverage.

\begin{table}[htbp]
    \centering
    \caption{\textbf{Ablation study on prototype refinement strategies.}}
    \label{tab:ablation_refinement}
    \begingroup
    \footnotesize
    \setlength{\tabcolsep}{1pt}
    \renewcommand{\arraystretch}{1.18}
    \begin{tabular*}{\linewidth}{@{\extracolsep{\fill}}l*{7}{c}@{}}
        \toprule
        \multirow{2}{*}{\textbf{Setting}} & \multicolumn{2}{c}{\textbf{ScanRef}} & \multicolumn{2}{c}{\textbf{ScanCap}} & \multicolumn{2}{c}{\textbf{ScanQA}} & \textbf{Avg.} \\
        \cmidrule(lr){2-3}\cmidrule(lr){4-5}\cmidrule(lr){6-7}
        & Acc@0.25 & Acc@0.5 & B-4@0.5 & C@0.5 & CIDEr & EM & (\%) \\
        \midrule
        w/o Refine. & 56.22 & 49.87 & 37.37 & 66.70 & 94.11 & 28.15 & 91.40 \\
        w/o Const. & 56.25 & 49.90 & 37.60 & 65.39 & 94.03 & 28.21 & 91.27 \\
        \midrule
        \textbf{w/ All} & \textbf{56.36} & \textbf{49.95} & \textbf{37.71} & \textbf{66.82} & \textbf{95.04} & \textbf{28.32} & \textbf{91.87} \\
        \bottomrule
    \end{tabular*}
    \endgroup
\end{table}
\endgroup

\begin{figure*}[t]
    \centering
    \captionsetup{labelfont=bf}
    \includegraphics[width=\linewidth]{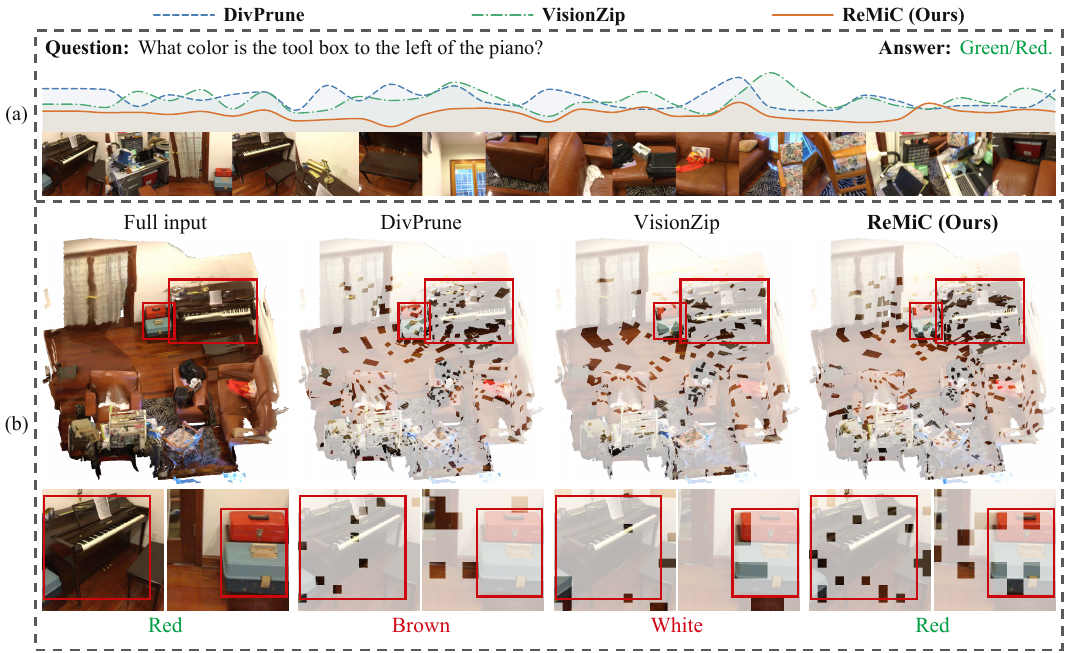}
    \caption{\textbf{Qualitative comparison.}
        (a) Frame-wise maximum representation errors of different pruning methods.
        (b) Visualization of retained tokens in the 3D scene.
        Red boxes highlight the piano and both toolboxes.}
    \label{fig:toolbox_case}
\end{figure*}

\subsection{Qualitative Results}
\label{sec:visualization}

As shown in Fig.~\ref{fig:toolbox_case}(a), ReMiC maintains consistently low maximum representation errors with small fluctuations across frames, whereas DivPrune and VisionZip exhibit higher errors and pronounced variations.
The higher errors indicate that parts of the scene are poorly represented, while the large fluctuations reveal uneven representation quality across views, suggesting that the token budget favors certain regions while leaving others underrepresented.
Fig.~\ref{fig:toolbox_case}(b) further visualizes the spatial distribution of retained tokens.
VisionZip tends to concentrate the token budget on background and floor regions, missing critical local objects.
DivPrune retains scattered tokens across the scene but fails to preserve fine-grained details of small or structurally complex objects.
By comparison, ReMiC maintains broad spatial coverage while preserving more complete object structures and local details.
Together, these observations highlight the importance of selecting tokens based on how well they represent the original scene, rather than relying solely on salience or feature diversity.

\begin{table}[!t]
    \centering
    \caption{\textbf{Efficiency analysis of our ReMiC.}
    Times are reported in seconds and Memory in GB.}
    \label{tab:efficiency_scanqa}
    \begingroup
    \small
    \setlength{\tabcolsep}{1pt}
    \renewcommand{\arraystretch}{1.18}
    \begin{tabular*}{\linewidth}{@{\extracolsep{\fill}}lc*{5}{c}@{}}
        \toprule
        \multirow{2}{*}{\textbf{Method}} & \textbf{Retention} & \multirow{2}{*}{\textbf{Pruning}} & \multirow{2}{*}{\textbf{Prefill}} & \multirow{2}{*}{\textbf{E2E}} & \multirow{2}{*}{\textbf{Memory}} & \multirow{2}{*}{\textbf{EM}} \\
        & \textbf{Ratio {\boldmath$R$}} & & & & & \\
        \midrule
        Video-3D LLM & 100\% & \textemdash & 1.584 & 2.394 & 18.17 & 30.10 \\
        \midrule
        3DZip & \multirow{3}{*}{20\%} & 0.173 & \textbf{0.359} & 1.348 & 16.87 & 28.56 \\
        DivPrune & & 0.130 & 0.391 & \textbf{1.312} & \textbf{16.85} & 28.66 \\
        \textbf{ReMiC} & & \textbf{0.094} & 0.398 & 1.327 & \textbf{16.85} & \textbf{29.30} \\
        \midrule
        3DZip & \multirow{3}{*}{10\%} & 0.123 & 0.231 & 1.165 & 16.82 & 27.83 \\
        DivPrune & & \textbf{0.054} & \textbf{0.230} & \textbf{1.077} & 16.77 & 27.14 \\
        \textbf{ReMiC} & & 0.098 & 0.236 & 1.176 & \textbf{16.75} & \textbf{28.32} \\
        \bottomrule
    \end{tabular*}
    \endgroup
\end{table}

\subsection{Efficiency Analysis}
\label{sec:efficiency_analysis}

Table~\ref{tab:efficiency_scanqa} compares the efficiency and accuracy of ReMiC, 3DZip, and DivPrune on ScanQA using a single NVIDIA RTX 3090 GPU under identical settings.
Pruning, Prefill, E2E, and Memory denote token pruning overhead, prefill latency, end-to-end latency, and peak reserved GPU memory, respectively.
At 10\% token retention, ReMiC achieves $6.72\times$ and $2.04\times$ speedups in prefill and E2E inference, respectively, while reducing peak reserved GPU memory by 7.8\% compared with the base model.
Benefiting from parallel local tree construction, ReMiC requires 20.5\% less pruning time than 3DZip while improving EM by 0.49 points.
Notably, while the pruning overhead of both baselines varies with the retention budget, ReMiC maintains nearly constant overhead, enabling a flexible trade-off between efficiency and accuracy.
\FloatBarrier
\section{Conclusion}
\label{sec:conclusion}

We introduce ReMiC, a training-free, representation-driven token pruning framework for 3D VLMs based on minimax hierarchical clustering.
ReMiC combines local prototype initialization for adaptive token allocation across spatial regions with global prototype refinement to incorporate cross-region information.
By explicitly modeling the representation gap between the original and retained token sets, ReMiC preserves scene-level coverage and fine-grained details.
Experiments across three 3D VLMs and five 3D scene understanding benchmarks demonstrate that ReMiC consistently outperforms existing pruning methods while substantially accelerating inference.
\section*{LLM Usage Statement}

The authors used OpenAI tools (ChatGPT/GPT) to assist with editing and refining manuscript text, related work and citations, as well as figures and diagrams.
These tools also supported the refinement of experimental designs, procedures, result analyses, research ideas, hypotheses, claims, and conclusions developed by the authors.
Additionally, the tools were used to generate code and provide reviewer-style feedback on the manuscript.
All AI-generated content incorporated into the work was reviewed, edited, or approved by the authors.
No large language models (LLMs) or other generative AI tools were used for synthetic data generation, dataset creation, or annotation.
The authors assume full responsibility for the accuracy, originality, integrity, and overall content of the submitted work.

\end{document}